\pdfoutput=1

\documentclass[11pt]{article}

\usepackage{acl}

\usepackage{times}
\usepackage{latexsym}
\usepackage{graphicx}
\usepackage{multirow}  
\usepackage{todonotes}

\usepackage[T1]{fontenc}

\usepackage[utf8]{inputenc}

\usepackage{microtype}

\title{Agreeing to Disagree: Annotating Offensive Language Datasets \\with Annotators' Disagreement}

  \author{Elisa Leonardelli, Stefano Menini, Alessio Palmero Aprosio\\
\textbf{Marco Guerini, Sara Tonelli} \\
 Fondazione Bruno Kessler, Trento, Italy \\
  {\tt \{eleonardelli,menini,aprosio,guerini,satonelli\}@fbk.eu} \\  }

\begin{document}
\maketitle
\begin{abstract}
Since state-of-the-art approaches to offensive language detection rely on supervised learning, 
it is crucial to quickly adapt them to the continuously evolving scenario of social media. While several approaches have been proposed to tackle the problem from an algorithmic 
perspective, so to reduce the need for annotated data, less attention has been paid to the quality of these data. Following a trend that has emerged recently, we focus on the level of agreement among annotators while selecting data to create offensive language datasets, a task involving a high level of subjectivity. 
Our study comprises the creation of three novel datasets of English tweets covering different topics and having five crowd-sourced judgments each. We also present an extensive set of experiments showing that  selecting training and test data according to different levels of annotators' agreement
has a strong effect on classifiers performance and robustness.  
Our findings are further validated in cross-domain experiments and studied using a popular benchmark dataset. We show that such hard cases, where low agreement is present, are not necessarily due to poor-quality annotation and we advocate for a higher presence of ambiguous cases in future datasets, particularly in test sets, to better account for the different points of view expressed online.
\end{abstract}

\section{Introduction}
\label{sec:intro}
When creating benchmarks for NLP tasks through crowd-sourcing platforms, it is important to consider possible issues with inter-annotator agreement. Indeed, crowd-workers do not necessarily have a linguistic background and are not trained to perform complex tasks, thus jeopardizing benchmark quality. Furthermore, some crowd-workers try to maximize their pay by supplying quick answers that have nothing to do with the correct label. This issue has been tackled in the past by proposing approaches to control for annotators' expertise and reliability \cite{hovy-etal-2013-learning}, trying to identify spammers and mitigate their effect on annotation, or by repeating labeling on targeted examples \cite{10.1145/1401890.1401965}.
However, not all tasks are the same: while in some cases, like for instance PoS-tagging or parsing, disagreement among annotators is more likely due to unclear annotation guidelines and can usually be reconciled through adjudication, full annotators' agreement should not be necessarily enforced in \textit{social computing tasks}, whose goal is to study and manage social
behavior and organizational dynamics, especially in virtual worlds built over the
Internet  \cite{4338496}. In these tasks -- which include offensive language detection among others -- subjectivity, bias and text ambiguity play an important role 
\cite{10.1145/3308560.3317083}, and being an inherent component of the task they should be measured and analysed rather than discarded \cite{DBLP:conf/swisstext/KlennerGA20,DBLP:conf/aiia/Basile20}. Indeed, instead of  aiming for a global consensus on what constitutes verbal abuse on social media, we investigate the impact of different degrees of disagreement, how classifiers behave with ambiguous training and test data, and the role of disagreement in current shared tasks. 
More specifically, we first collect and annotate three datasets of English tweets covering different domains, to test if agreement among a pool of generic classifiers can be considered a proxy for annotator agreement. 
We then focus on how annotator agreement (both in training and test set) impacts classifiers' performance, considering domain-specific and generic classifiers as well as in-domain and out-of-domain experiments. We also show that low agreement examples -- no matter how difficult they can be -- still provide useful signal for training offensive language detection systems and do not represent random annotations. So ``coin-flipping'' or example removal seems not to be the right strategy to solve these disagreement cases.
Then, we measure disagreement in the English test set of the last Offenseval shared task \cite{zampieri-etal-2020-semeval}, and analyse to what extent the high performance achieved by most participating systems is related to high agreement in annotation.

We release the new annotated datasets upon request,\footnote{See Ethics Statement section for further details} including more than 10k tweets covering three domains. The messages have been labeled with 50k crowd-worker judgements and annotated with agreement levels. To our knowledge, this represents the first dataset explicitly created to cover different agreement levels in a balanced way.
We also advocate for the release of more datasets like the one we propose, especially for highly subjective tasks, where the need to include different points of view should be accounted for.

\textit{NOTE}:  This  paper  contains  examples  of  language  which  may  be  offensive  to  some  readers. They do not represent the views of the authors.

\section{Related Work}
\label{sec:relwork}
While there has been an extensive discussion on minimal standards for inter-annotator agreement to ensure data quality
\cite{di-eugenio-glass-2004-squibs,passonneau-2004-computing,artstein-poesio-2008-survey}, recently an increasing number of works argue that disagreement is unavoidable because language is inherently ambiguous \cite{Aroyo_Welty_2015}, proposing ways to tackle annotators' disagreement when building training sets \cite{dumitrache-etal-2019-crowdsourced}. \newcite{hsueh-etal-2009-data}, for example, identify a set of criteria to select informative yet unambiguous examples for predictive modeling in a sentiment classification task. \newcite{rehbein-ruppenhofer-2011-evaluating} analyse the impact that annotation noise can have on active learning approaches. Other works along this line investigate the impact of uncertain or difficult instances on supervised classification \cite{9010969}, while \newcite{beigman-klebanov-beigman-2014-difficult}  
show that including hard cases in training data results in poorer classification of easy data in a word classification task.  Along the same lines, \newcite{jamison-gurevych-2015-noise} show that filtering instances with low agreement improve classifier performance in four out of five tasks. Both works observe that the presence of such instances lead to misclassifications. 

Several approaches have been presented that implement strategies to deal with disagreement when training classifiers for diverse tasks. In most cases, disagreement has been treated as a consequence of low annotation quality, and addressed through methodologies aimed at minimising the effects of noisy crowdsourced data. \newcite{Simpson_Pfeiffer_Gurevych_2020}, for example,  present a Bayesian sequence combination approach to train a model directly from crowdsourced labels rather than aggregating them. They test their approach on tasks such as NER where disagreement is mainly due to poor annotation quality. Other works have focused instead on uncertainty in PoS-tagging, integrating annotators' agreement in the modified loss function of a structured perceptron \cite{plank-etal-2014-learning}. Also \newcite{DBLP:conf/aaai/RodriguesP18} propose an approach to automatically distinguish the good and the unreliable annotators and capture their individual biases. They propose a novel crowd layer in deep learning classifiers to train neural networks directly from the noisy labels of multiple annotators, using only backpropagation. 

Other researchers have suggested to remove hard cases from the training set \cite{10.1162/coli.2009.35.4.35402} because they may potentially lead to poor classification of easy cases in the test set. We argue instead that disagreement is inherent to the kind of task we are going to address (i.e. offensive language detection) and, in line with recent works, we advocate against forced harmonisation of annotators' judgements for tasks involving high levels of subjectivity \cite{DBLP:conf/swisstext/KlennerGA20,DBLP:conf/aiia/Basile20}. Among recent proposals to embrace the uncertainty exhibited by human annotators, \newcite{10.1145/3411764.3445423} propose a novel metric to evaluate social computing tasks that disentangles stable opinions from noise in crowd-sourced datasets. \newcite{Akhtar_Basile_Patti_2020}, instead, divide the annotators into groups based on their polarization, so that different gold standard datasets are compiled and each used to train a different classifier. 

Compared to existing works, our contribution is different in that we are interested mainly in the process of dataset creation rather in evaluation metrics or classification strategies. Indeed, our research is guided mainly by research questions concerning the data selection process, the composition of datasets and the evaluation using controlled levels of agreement. To this purpose, we create the first dataset for offensive language detection with three levels of agreement and balanced classes, encompassing three domains. This allows us to run comparative in-domain and out-of-domain evaluations, as well as to analyse existing benchmarks like the Offenseval dataset \cite{zampieri-etal-2020-semeval} using the same approach. While few crowd-sourced datasets for toxic and abusive language detection have been released with disaggregated labels \cite{hateoffensive}, they have not been created with the goal of analysing  disagreement, therefore no attention has been paid to balance the number of judgments across different dimensions, like in our case.

\section{Data Selection and Annotation}
\label{sec:data}

In our study, we focus on three different domains, which have been very popular in  online conversations in 2020: Covid-19, US Presidential elections and Black Lives Matter (BLM) movement.  After an empirical analysis of online discussions, a set of hashtags and keywords for each domain are defined (e.g. \textit{\#covid19}, \textit{\#election202}, \textit{\#blm}). Then, using Twitter public APIs, tweets in English containing at least one of the above keywords  are collected in a time span between January and November 2020 (for more details about data collection see Appendix \ref{appendixd}). From this data collection, we randomly select 400,000 tweets (around 130,000 for each domain), which we then pre-process by splitting hashtags into  words using the Ekphrasis tool \cite{gimpel2010part} and then replacing all mentions to users and urls with $\left<\mathrm{user}\right>$ and $\left<\mathrm{url}\right>$ respectively.

\subsection{Ensemble of classifiers to select data for annotation}
\label{sub:ensemble}

Since we do not know  the real distribution of agreement levels in the data we collected, random sampling for annotation might be a sub-optimal choice. Thus, we developed a strategy to pre-evaluate the tweets, trying to optimize annotators' effort by having a balanced dataset (in fact data might be very skewed leading to over-annotation of some classes and under-annotation of others). 
To pre-evaluate the tweets we use a heuristic approach by creating an ensemble of 5 different classifiers, all based on the same BERT configuration and fine-tuned starting from the same abusive language dataset \cite{founta2018large}. Since the original dataset contains four classes (\texttt{Spam}, \texttt{Normal}, \texttt{Abusive} and \texttt{Hateful}), we first remove the tweets from the \texttt{Spam} class and map the remaining ones into a binary \texttt{offensive} or \texttt{non-offensive} label, 
by merging \texttt{Abusive} and \texttt{Hateful} tweets into the \texttt{offensive} class and mapping the \texttt{Normal} class into the \texttt{non-offensive} one. We then select 15k tweets from the Founta dataset (\textasciitilde 100k tweets) for speeding up the process, as we are not interested in the overall performance of the different classifiers, but rather in their relative performances.
Each classifier of the ensemble is trained using a different balance for the training and the evaluation set, so to yield slightly different predictions. 
In particular, all five classifiers are trained with the BERT-Base uncased model\footnote{12-layer, 768-hidden, 12-heads, 110M parameters \url{https://github.com/google-research/bert}}, a max seq length of 64, a batch size of 16 and 15 epochs. 
One classifier has been trained using 12k tweets in the training and 3k in the validation set, a second classifier was trained using the same training instances but repeated twice (24k), while the validation set remained the same. In a third and fourth configuration, we repeat twice the offensive and the non-offensive training instances respectively. Finally, in a fifth configuration we change the proportion between training and validation set (10k for training, 5k for validation).

The rationale for this choice is twofold: (i) since we will collect 5 crowd-annotations for each tweet, we want to have an intuitive and possible direct comparison between ensemble agreement and annotators' agreement (i.e. five votes per tweet coming from the classifiers and five from crowd-workers).  (ii) The dataset in \newcite{founta2018large} has been specifically created to encompass several types of offensive language. We can therefore consider it as a  general prior knowledge about verbal abuse online before adapting our systems to the 3 domains of interest.

In the following sections we will denote \textit{unanimous agreement} with $A^{++}$ (i.e. agreement between 5 annotators or classifiers), \textit{mild agreement} with $A^+$ (i.e. 4 out of 5 annotations agreeing on the same label), and \textit{weak agreement} with $A^0$ (i.e. the 5 annotations include 3 of them in agreement and 2 in disagreement). When focusing also on the label we will use the same notation, representing offensive tweets as $O^{++/+/0}$ and non offensive ones as $N^{++/+/0}$ respectively.

The pre-evaluation through the classifier ensemble resulted in the following agreement distribution: about 92\% of the data was classified as $A^{++}$. For about 5\% of the data, agreement among the  classifiers was $A^+$, while for the remaining 3\% of the data, they fell in the $A^0$ situation.

\subsection{Data Annotation with AMT}

In order to analyse the relation between automated and manual annotation with respect to agreement and disagreement, we select an equal number of tweets from each class of agreement of the ensemble ($A^{++}$, $A^{+}$, $A^{0}$) to be manually annotated. For each domain and each agreement class we select 1,300 tweets -- equally divided between \texttt{offensive} and \texttt{non-offensive} predictions -- for a total of 3,900 tweets per domain.

Every tweet is annotated by 5 native speakers from the US, who we expect to be familiar with the topics, using Amazon Mechanical Turk. We follow for all domains the same annotation guidelines, aimed at collecting crowd-workers' judgements on the offensiveness of the messages using the binary labels \texttt{offensive} and  \texttt{not offensive} (see Guidelines included in Appendix \ref{appenda}).

To ensure high-quality annotations, we select a pool of tweets from the three domains of interest and ask three expert linguists to annotate them. The tweets with perfect agreement are used as gold standard. We then include a gold standard tweet in every HIT (group of 5 tweets to be annotated). If a crowd-worker fails to evaluate the gold tweet, the HIT is discarded.
Moreover, after the task completion we remove all the annotations done by workers who did not reach a minimum overall accuracy of 70\% with respect to the gold standard.
As a consequence of this quality control, for some tweets we could not collect five annotations, and they had to be removed from the final dataset. On the other hand, it was a crucial process to minimise the possible impact of spam and low-quality annotations on disagreement -- which is the focus of our analysis. 
The total number of tweets annotated using AMT is 10,753, including 3,472 for Covid-19, 3,490 for US elections and 3,791 for BLM.
Some (slightly modified) examples of tweets judged with different levels of agreement by crowd-annotators are reported in Table \ref{my_agrees}.

\begin{table*}[h]
    \centering
    \scalebox{0.95}{
    \begin{tabular}{|l|l|}
    \hline
\multirow{2}{*}{$N^{++}$} & Stand for something or else fall for anything. \#BlackLivesMatter  \\
&   Hello world! What a great day to be alive \#Trump2020 \#MAGA \\
\hline
\multirow{2}{*}{$N^{+}$} &Come on man! Lock’em up!!! \#maga\\
&    Not the first time. You all misspelled \#blacklivesmatter. Speak up! @user  \\
\hline
\multirow{2}{*}{$N^{0}$}  & Set fire to Fox News (metaphorically) \\
&  @user  is outing \#BLACK\_LIVES\_MATTER as a cult! HE IS CORRECT!\\
\hline
\multirow{2}{*}{$O^{0}$} &  \#DISGUSTING \#Democrats terrorize old folks just before \#elections2020 \\
& I love this shit! \#BlackLivesMatter \\
\hline
\multirow{2}{*}{$O^{+}$} &  @user You're a bumbling fool \#elections2020 \\

& Elections 2020: Red Rapist v. Blue Racist \\
\hline

\multirow{2}{*}{$O^{++}$} &   Y'all trending about kpop stans instead of \#BlackLivesMatter big fack you \\
&  Crazy idiots. This is batshit bullshit. \#elections2020 \\   
    \hline 
    \end{tabular}}
    \caption{Examples of tweets with different degrees of crowd-workers' agreement. The messages have been created starting from real examples by slightly changing their wording, so to make it impossible to retrieve the original ones on Twitter. $N$=Not offensive, $O$=Offensive. $++$/$+$/$0$ correspond to high, medium and low agreement respectively.}
    \label{my_agrees}
\end{table*}

\subsection{Annotators and Ensemble Agreement}

\begin{figure}[h]
    \centering
    \includegraphics[width=7.8cm]{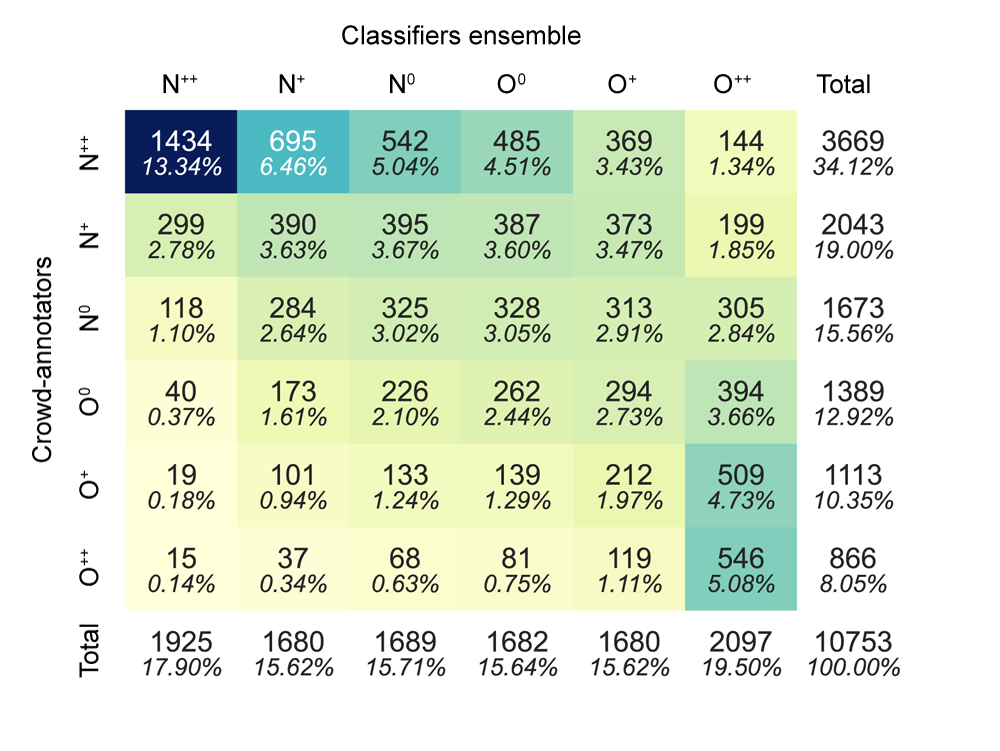}
    \caption{Confusion matrix (raw number of tweets and percentage) between classifiers ensemble agreement (x-axis) and crowd-annotators agreement (y-axis) on offensive (``O'') and non-offensive (``N'') labels. 
    }
    \label{fig:galaxy}
\end{figure}

If we use the majority vote for crowd-annotated data, the datasets have an average distribution of 31\% of offensive and 69\% non-offensive tweets, while it is 50\% each according to ensemble annotation we used for sampling. 
This means that our classifiers tend to label more tweets as offensive compared to human annotators, as shown in the confusion matrix in Fig. \ref{fig:galaxy}. It is interesting to note that, although the tweets to be annotated were selected evenly across classifiers' agreement classes, the agreement between annotators is not uniformly distributed.

As regards annotators' agreement, for about 43\% of the tweets annotated we have full consensus between annotators ($A^{++}$). The vast majority of these tweets were judged unanimously as non-offensive (34,12\% $N^{++}$), and only 8,05\% of the data were judged unanimously offensive ($O^{++}$), the less represented type of agreement. For the remaining data, 29,35\% has mild agreement ($A^+$, 4 out of 5 annotators agreed)  with 19\% $N^{+}$ and 10,35\% $O^{+}$, and another 28,28\% of the data in the class $A^0$  (3 vs 2 annotators) with 15,56\% $N^{0}$ and 12,92\% $O^{0}$.

We also compute Pearsons' correlation coefficient between the agreement of the ensemble classifiers and that of annotators. It achieves a moderate correlation ($r$ = 0.51), showing that training an ensemble of classifiers on generic data to pre-screen domain-specific tweets before manual annotation could help identifying tweets that are either unambiguous or more challenging. A similar correlation ($r$ = 0.50) was obtained on an ensemble of BiLSTM classifiers trained with the same training and development sets of the five BERT-based classifiers, suggesting that the pre-screening approach could be used also  with other classifiers.

\subsection{Qualitative analysis of (dis)agreement}

Through a manual analysis of the tweets belonging to the $A^0$ class, we can identify few phenomena that lead to disagreement in annotation. In many cases, tweets are ambiguous and more context would be needed to fully understand whether the user wanted to offend someone or not.  These cases include the presence of deictic expressions or pronouns referring to previous tweets, see for example:

\begin{quote}
(1) \textit{Shoulda thrown this clowns bike off the bridge! }
\\
(2) \textit{Won't work.  Gangs will terrorize the city.  Murder at will and maybe they'll shoot the Mayor.}
\end{quote}

Other cases include generic expressions of anger that are not targeted against a specific person or group, or expressions of negative feelings, see for example:

\begin{quote}
(3) \textit{Amen ! Enough of this crap !}
\end{quote}

Finally, questions, and in particular rhetorical questions, are very frequent in the $A^0$ class  and their interpretation seems to represent a challenging 
task for crowd-workers:

\begin{quote}
(4) \textit{if George Floyd was white would the cop have acted in the same violent, murderous way?}
\\
(5) \textit{What is it with these kids of leftist politicians?}
\end{quote}

Overall, disagreement does not seem to stem from poor annotation of some crowd-workers, but rather from genuine differences in the interpretation of the tweets. Additionally, BLM and US American elections are recent events and annotators may have been biased by their personal opinion on the topic during  annotation, an effect that has already been highlighted in  \newcite{sap-etal-2019-risk,DBLP:conf/acl/SapGQJSC20}.

\section{Classification experiments}
\label{sec:exp}

After collecting information on human agreement on tweets covering three different domains, we aim at assessing the impact of (dis)agreement on classifier behaviour.

To this end, we  create several balanced configurations of the datasets, so to control for the effect of agreement level, label distribution and domain topic. 
We first split the data into a training and test set of 75\% and 25\% for each domain. Then, to control for the effect of training data size, we further downsample all sets to the smallest one, so that each agreement sample is equally represented ($A^{++}$, $A^{+}$, $A^{0}$).  In this way, we obtain 3 sets of training data -- one per ambiguity level --  containing 900 tweets each. Every set further contains 300 tweets from each domain, half for \texttt{offensive} label and half for \texttt{non-offensive} label so to control also for the effect of label distribution across domains and agreement levels. 

\subsection{Impact of (dis)agreement in training data}
\label{subsec:2}

To assess the impact of agreement level in training data, we run a series of experiments by comparing two different classifiers: the first one relies on BERT directly fine-tuned on domain data, while the second foresees also an intermediate fine-tuning step using the entire dataset in \newcite{founta2018large}, inspired by the supplementary training approach from \newcite{phang2018sentence}. BERT is used with the same parameters of the ensemble classifiers, reported in Section \ref{sub:ensemble}.
The domain data used for fine-tuning are built starting from the training data described above divided into different agreement levels ($A^{++}$, $A^{+}$, $A^{0}$ and their combinations).
 
 \begin{table}[h]
    \centering
    \small
    \scalebox{1}{
    \begin{tabular}{|l|r|cc|}
    \hline 
Training&\multicolumn{1}{|l|}{Training}&\multirow{2}{*}{All domains}&\multicolumn{1}{l|}{Founta +}\\
 split & \multicolumn{1}{|l|}{size}    	&			&	 \multicolumn{1}{l|}{all domains} 	\\
      
\hline
$A^{++}$	    & 900	&	0.746		&	\textbf{0.757}	\\
$A^+$	    & 900	&	0.734		&	0.753	\\
$A^0$	    &900	&	0.639		&	0.683	\\

\hline
$A^{++/+}$	& 1800	&	\textbf{0.755}		&	\textbf{0.756}	\\
$A^{+/0}$	& 1800	&	0.728		&	0.724	\\

$A^{++/0}$	& 1800	&	0.723	&	0.730	\\

\hline
$A^{++/+/0}$	    & 2700  	&	0.745	&	0.752	\\

			\hline
    \multicolumn{4}{|l|}{\rule{0pt}{3ex}Baseline - training only on Founta et al. data: 0.667 F1}\\
\hline 
    \end{tabular}}
    \caption{Performance (F1) when training on data with different levels of human agreement (rows), fine-tuned either on domain data or using the dataset from \newcite{founta2018large} and domain data.}
    \label{tab:f1-agreement}
\end{table}

Results are reported in Table \ref{tab:f1-agreement}. Note that, for training, the tweets in a given partition  for all domains are merged, while they are tested on each domain separately. The reported F1 is an average of the three results (results for each domain can be found in the Appendix and are consistent with the ones reported here). We observe that, if we consider only one level of agreement, data with total agreement are the best for prediction ($A^{++}$), up to the point that $A^{++}$ data alone provide better results than using all data available in the three splits (\textit{all}), despite the different size (900 vs. 2700 instances). 
Additionally, the combination of high and mild agreement data ($A^{++/+}$) yields results that are in line with the best configuration obtained with two fine-tuning steps (0.755 vs 0.757). This result clearly indicates that for this kind of task it is not necessary to collect huge datasets for fine-tuning, since few data from the target domain may suffice if properly selected. Finally, the effect of using low agreement data for training is detrimental, in line with findings reported in past works \cite{reidsma-op-den-akker-2008-exploiting, jamison-gurevych-2015-noise}. This can be spotted in two results: the use of generic data alone as in our baseline is better than using low agreement in-domain data (0.667 vs. 0.639) and all configurations where $A^0$ is added to mild and high agreement data perform worse than without $A^0$ (0.734 vs 0.728 and 0.746 vs 0.723).

\subsection{Impact of (dis)agreement in test data}

As a next step, we investigate how classifier's performance varies as a function of annotators' agreement in the test data. 
To this end, we divide also our test set into subsets according to the same agreement levels ($A^{++}$, $A^{+}$, $A^{0}$) and calculate separate F1s on each of these splits. We run the classifier for `all domains' described in Section \ref{subsec:2}, i.e. trained on the three domains and tested on one of them. 
Results, reported in Table \ref{tab:f1-subclasses}, are obtained by averaging the F1 for each domain.

We observe a dramatic drop in performance when agreement decreases in the test set, indicating that ambiguous data are the most challenging to classify. These results highlight the need to control for ambiguity also in the test set when creating offensive language benchmarks (for example in shared tasks), in order to avoid high system performance being due to a lack of challenging examples. 
The best performance on ambiguous data is obtained when training on unambiguous and mildly ambiguous data ($A^{++/+}$). Interestingly, adding $A^+$ data to $A^{++}$ data leads to the highest increase in performance exactly for $A^0$ test data (from 0.552 to 0.574). This rules out the possibility that a certain level of disagreement in the training set is more effective in classifying the same type of ambiguity in the test set (e.g. train and test on $A^0$ data), and suggests that high agreement or mild  agreement training sets perform better in all cases.

\begin{table}[h]
    \centering
    \small
    \scalebox{1}{
    \begin{tabular}{|l|r|l|c|}
    \hline 
Training split    & Training size & Tested on	&	F1	\\
\hline 
$A^{++/ +}$ & 1800  & $A^{++}$	&	0.860	\\
$A^{++/ +}$ &  1800 & $A^+$	&	0.768\\	
$A^{++/ +}$ &  1800 & $A^0$	&	0.574\\	
\hline 
$A^{++}$	 &   900 & $A^{++}$	&	0.847	\\
$A^{++}$	  &	900 & $A^+$	&	0.763	\\
$A^{++}$	  &	900 & $A^0$	&	0.552	\\
\hline 
$A^0$	    & 900 & $A^{++}$	&	0.662	\\
$A^0$	    & 900 & $A^+$	&	0.639	\\
$A^0$	    & 900& $A^0$	&	0.567	\\
\hline 
    \end{tabular}}
    \caption{Performance on  $A^{++/ +}$, $A^{++}$, $A^0$ data, classified with ``all domains" configuration in Table \ref{tab:f1-agreement}. 
    }
    \label{tab:f1-subclasses}
\end{table}

\subsection{Impact of (dis)agreement on out-of-domain data}
We then test the effect of cross-domain classification according to agreement levels, so to minimise the impact of possible in-domain overfitting. We repeat the experiments described in the previous section by using two domains for training and the third for testing. As an example, a classifier model was trained using $A^{++}$ data from Covid 19 and US Presidential campaign, and tested on $A^{++}$ data on BLM. This has been repeated for each domain and each agreement level. For conciseness of presentation, we report in Table \ref{tab:performance} the F1 obtained by averaging F1 on each domain (results for each domain can be found in the Appendix and also in this case  they are consistent with the one reported here). Results confirm  that (i) the classifier yields a good performance when the training data have high agreement, even in an out-of-domain scenario, and (ii) adding $A^0$ data to the training set has a detrimental effect on performance. Finally, if we compare these results with Table     \ref{tab:f1-agreement}, we observe that the effect of overfitting on in-domain data is very limited.

\begin{table}[h]
    \centering
    \small
    \scalebox{1}{
    \begin{tabular}{|l|r|cc|}
\hline 
Training  & Training &	\multicolumn{1}{l}{Out of} 	&	\multicolumn{1}{l|}{Founta + out}  \\
split   &   \multicolumn{1}{|l|}{size}   &domain& \multicolumn{1}{l|}{of domain}\\ \hline
$A^{++}$	& 600 &	0.719	&	0.747	\\
$A^+$ &600 & 0.677 &0.716\\
$A^0$ &600&0.567 & 0.658\\
\hline
$A^{++ / +}$ & 1,200 &	\textbf{0.732}	&	\textbf{0.748}	\\
$A^{+/0}$	& 1,200	&	0.659	& 0.715	\\
$A^{++/0}$	& 1,200	&	0.714	& 0.722	\\
\hline
$A^{++/+/0}$	& 1,800 &	0.722	&	0.737	\\  
\hline 
\end{tabular}}
\caption{Performance (F1) in the out-of-domain setting. 
Results are the average F1 obtained by the classifier on each domain when trained on the other two. }
\label{tab:performance}
\end{table}

\subsection{(Dis)agreement versus Randomness}

An additional question we want to address is whether low agreement data provide some useful information for training offensive language detection systems or if the effect of such data is no more that of random annotation.

We therefore replicate the experiments of Table \ref{tab:f1-agreement} by replacing the label of $A^0$ data with a random one. Since we want to obtain the same controlled distribution we assign the same probability to $N$ and $O$ labels. Results are reported in Table \ref{tab:f1-rand-agreement}. As can be seen, when using $A^{0_{rand}}$ data the results worsen as compared to $A^0$, indicating that the label in $A^0$ are not assigned by chance and they can contain useful signal for the classifier, albeit challenging. Consistently with previous results, the more gold and high agreement data is added to the training, the smaller the effect of $A^{0_{rand}}$. These results show also that coin-flipping, which has been suggested in past works to resolve hard disagreement cases \cite{10.1162/coli.2009.35.4.35402}, may not be ideal because it leads to a loss of information.

 \begin{table}[h]
    \centering
    \small
    \scalebox{1}{
    \begin{tabular}{|l|r|cc|}
    \hline 
Training&\multicolumn{1}{|l|}{Training}&\multirow{2}{*}{All domains}&\multicolumn{1}{l|}{Founta +}\\
 split & \multicolumn{1}{|l|}{size}    	&			&	 \multicolumn{1}{l|}{all domains} 	\\
      
\hline
$A^0$	    &900	&	0.639		&	\textbf{0.683}	\\
$A^{0_{rand}}$	    &900	&	0.505		& 0.576	\\
\hline
$A^{+/0}$	& 1800	&	\textbf{0.728}		&	0.724	\\
$A^{+/0_{rand}}$	& 1800	&	0.657		& 0.689	\\
\hline
$A^{++/0}$	& 1800	&	0.723	&	\textbf{0.730}	\\
$A^{++/0_{rand}}$	& 1800	&	0.684		& 0.703	\\
\hline
$A^{++/+/0}$	    & 2700  	&	0.745	&	\textbf{0.752}	\\
$A^{++/+/0_{rand}}$	    & 2700  	&	0.719	&	0.730	\\
\hline 
    \end{tabular}}
    \caption{Performance (F1) when training on data with different levels of human agreement (rows) and replacing $A^0$ labels with random ones ($A^{0_{rand}}$). First line of each group is reported from Table~\ref{tab:f1-agreement} for comparison.}
    \label{tab:f1-rand-agreement}
\end{table}

\section{Experiments on Offenseval dataset}

Our experiments show that when training and test data include tweets with different agreement levels, classification of offensive  language is still a challenging task. Indeed, our classification results reported in Table \ref{tab:f1-agreement} and \ref{tab:performance} suggest that on this kind of balanced data, F1 with Transformer-based models is $\approx$0.75. However, system results reported for the last Offenseval shared task on  offensive language identification in English tweets \cite{zampieri-etal-2020-semeval} show that the majority of submissions achieved an F1 score > 0.90 on the binary classification task. 

We hypothesize that this delta in performance may depend on a limited presence of low agreement instances in the Offenseval 
dataset used for evaluation \cite{zampierietal2019}. 
We therefore randomly sample 1,173 tweets from the task test data (30\% of the test set) and annotate them with Amazon Mechanical Turk using the same process described in the previous sections (5 annotations per tweet). We slightly modify our annotation guidelines by including the cases of profanities, which were explicitly considered offensive in Offenseval guidelines.

Results, reported in Table \ref{tab:offenseval-comparison} (left column) show that the outcome of the annotation is clear-cut: more than 90\% of the tweets in the sample have either a high ($A^{+}$) or very high ($A^{++}$) agreement level. Furthermore, only 6.4\% of the annotations (75) have a different label from the original Offenseval dataset, 50\% of which are accounted for by the $A^0$ class alone. So our annotation is very consistent with the official one and the distribution is very skewed towards high agreement levels, as initially hypothesized.

\begin{table}[h]
    \centering
    \small
    \scalebox{1}{
    \begin{tabular}{|l|rr|rr|}
    \hline 
Agreement & \multicolumn{2}{c|}{Offenseval}& \multicolumn{2}{c|}{400k Tweets}\\
\hline 
 $A^{++}$ & 75.62\% &(887) & 68.52\% &(274,514) \\
 $A^{+}$ & 14.75\% &(173) & 19.08\%& (76,457)\\ 
 $A^0$ & 9.63\% &(113) & 12.40\% &(49,694)\\
\hline  
 $N^{++}$& 64.36\% &(755)  &65.12\% &(260,925) \\
 $N^{+}$& 5.80\% &(68) &15.50\% &(62,085) \\
 $N^0$& 4.60\% &(54) &7.94\% &(31,813) \\
 $O^0$& 5.03\% &(59) &4.46\% &(17,882) \\
 $O^{+}$& 8.95\% &(105) &3.59\%& (14,372) \\
 $O^{++}$& 11.25\% &(132) &3.39\% &(13,589) \\
 \hline
\end{tabular}}
\caption{Comparison of agreement distribution in Offenseval sample and projection on 400k tweets.}
\label{tab:offenseval-comparison}
\end{table}

To understand whether this skewness can be generalised, i.e. if this sample distribution might be representative of a population distribution, we also estimate the distribution of agreement levels in the initial pool of data (around 400k tweets) we collected using US Election, BLM and Covid-related hashtags (Section \ref{sec:data}).\footnote{Since we pre-selected the tweets to be annotated through the classifier ensemble,
to estimate the real distribution of agreement levels in our data we classified with the ensemble all of them (400k tweets). Then, to determine the proportion of each class of agreement, we projected the distribution of annotators' agreement level for each ensemble class, using the confusion matrix reported in Figure \ref{fig:galaxy}.} The estimate of the distribution for class  $A^{+}$, $A^{++}$ and $A^0$ is reported in Table \ref{tab:offenseval-comparison} (right column).
A comparison between the two columns shows that 
disagreement distribution in the Offenseval sample is in line with the distribution in the data we initially collected before balancing, providing initial evidence that this distribution -- with few disagreement cases -- might be a `natural' one for online conversations on Twitter.

Differences emerge when considering the ratio of offensive tweets. In Offenseval data, the percentage of offensive tweets is more than double the percentage in our data (25.23\% vs. 11.44\%), because the authors adopted  several strategies to overrepresent  offensive tweets \cite{zampierietal2019}.

As a final analysis, we collect the runs submitted to Offenseval and compute the F1 score of each of these systems over the three levels of agreement separately. Overall, we consider all runs that in the task obtained F1 > 0.75, i.e. 81 runs out of 85. Results are reported in Table \ref{tab:f1-offenseval} as the average of the F1  obtained by the different systems. 
This last evaluation confirms our previous findings, since F1 increases when agreement level increases in test data. This finding, together with the distribution of agreement levels, shows that the high performance obtained by the best systems in the shared task is most probably influenced by the prevalence of tweets with total agreement. 

\begin{table}[h]
    \centering
    \small
    \scalebox{1}{
    \begin{tabular}{|l|r|r|}
    \hline 
Offenseval 2020 - test subsets     & F1 	& StDev \\
\hline 
 $A^{++}$  (887 tweets) &  0.915 & $\pm$   0.055  	\\
 $A^{+}$ (173 tweets)	&	0.817 & $\pm$   0.075	\\
 $A^{0}$ (113 tweets)	& 0.656 & $\pm$ 	 0.067 \\
\hline

    \end{tabular}}
    \caption{Average F1 obtained by the best systems at Offenseval 2020 $\pm$ StDev. }
    \label{tab:f1-offenseval}
\end{table}

\section{Discussion and Conclusions}

We have presented a data annotation process and a thorough set of experiments for assessing the effect of (dis)agreement in training and test data for offensive  language detection. We showed that an ensemble of classifiers can be employed to preliminarily select potentially unambiguous or challenging tweets. By analysing these tweets we found that they represent real cases of difficult decisions, deriving from interesting phenomena, and are usually not due to low-quality annotations. We also found that these challenging data are minimally present in a popular benchmark dataset, accounting for higher system performance. We believe that such hard cases should be more represented in benchmark datasets used for evaluation of hate speech detection systems, especially in the test sets, so to develop more robust systems and avoid overestimating classification performance. This goal can be achieved by integrating the common practice of oversampling the minority \textit{offensive} class with the oversampling of minority \textit{agreement} classes.

From a multilingual perspective, we also noted that at Offenseval 2020 the best performing systems on Arabic scored 0.90 F1 with a training set of 8k tweets, 0.85 on Greek with less than 9k tweets, and 0.82 on Turkish despite having more than 32k examples for training. This shows that the amount of training data is not sufficient to ensure good classification quality, and that also in this case a study on disagreement levels could partly explain these differences (this is further corroborated by the fact that for Turkish the lowest  overall inter-annotator agreement score was reported).

As future work, we plan to develop better approaches to classify (dis)agreement, in order to ease oversampling of low agreement classes. Preliminary experiments (not reported in this paper) show that the task is not trivial, since supervised learning with LMs such as BERT does not work properly when trying to discriminate between ambiguous and not ambiguous tweets. Indeed, BERT-based classification performed poorly both in the binary task (ambiguous vs. not ambiguous) and in the three-way one (offensive vs. not offensive vs. ambiguous). This suggests that ambiguity is a complex phenomenon where lexical, semantic and pragmatic aspects are involved, which are difficult to capture through a language model. 

This corpus, together with the experiments presented in this paper, will hopefully shed light onto the important role played by annotators' disagreement, something that we need to understand better and to see as a novel perspective on data. Indeed, if we want to include diversity in the process of data creation and reduce both the exclusion of minorities' voices and demographic misrepresentation \cite{hovy-spruit-2016-social}, disagreement should be seen as a signal and not as noise.

 \section{Ethics Statement}

The tweets in this dataset have been annotated by crowd-workers using Amazon Mechanical Turk. All requirements introduced by the platform for tasks containing adult content were implemented, for example adding a warning in the task title. We further avoid to put any constraints on the minimum length of sessions or on the minimum amount of data to be labeled by each crowd-worker, therefore they were not forced to prolonged exposure to offensive content. Indeed, we observed that crowd-workers tended to annotate for short sessions, on average 20 minutes, which suggests that annotating was not their main occupation. Crowd-workers were compensated on average with 6 US\$ per hour.

Although we put in place strict quality control during data collection, we compensated the completed hits also when annotations were finally discarded because they did not reach the minimum accuracy threshold of 70\% w.r.t. the gold standard. We also engaged in email conversations with crowd-workers when they were blocked because of mismatches with the gold standard tweets. In several cases, we clarified with them the issue and subsequently unlocked the task. 

Concerning the annotated dataset, we support scientific reproducibility and we would like to encourage other researchers to build upon our findings. However, we are aware that ethical issues may arise related to the complexity and delicacy of judgments of offensiveness in case they are made public. Therefore,  in compliance with Twitter policy, we want to make sure that our dataset  will be reused for non-commercial research only\footnote{https://developer.twitter.com/en/developer-terms/policy} avoiding any discriminatory purpose, event monitoring, profiling or targeting of individuals. The dataset, in the form of tweet IDs with accompanying annotation, can be obtained upon request following the process described at this link: \url{https://github.com/dhfbk/annotators-agreement-dataset}. 
Requestors will be asked to prove their compliance with Twitter policy concerning user protection and non-commercial purposes, as well as to declare that they will not use our dataset to collect any sensitive category of personal information. Also, releasing the tweet IDs instead of the text will enforce users' right to be forgotten, since it will make it impossible to retrieve tweets if their authors delete them or close their account.
Although we are aware of the risks related to developing and releasing hate speech datasets, this research was carried out with the goal of improving conversational health on social media, and even exposing the limitations of binary offensive language detection. We believe that our findings confirm the context- and perspective-dependent offensiveness of a message, and we therefore avoid binary labels, stressing the importance of taking multiple points of view (in our case, five raters) into account.
Following the same principle of avoiding profiling, crowd-workers' IDs are not  included in the dataset, so that it will not be possible to infer annotator-based preferences or biases.

\section*{Acknowledgments}

Part of this work has been funded by the  KID\_ACTIONS REC-AG project (n. 101005518) on ``Kick-off preventIng and responDing to children and AdolesCenT cyberbullyIng through innovative mOnitoring and educatioNal technologieS'', \url{https://www.kidactions.eu/}.

\bibliography{acl_latex}
\bibliographystyle{acl_natbib}

\appendix

\section{Annotation Guidelines for AMT}
\label{appenda}
This section contains the instructions provided to annotators on Amazon Mechanical Turk. The first part changes according to the domain:

\par{\textbf{Covid-19}: \textit{The tweets in this task have been collected during the pandemic. Would you find the content of the messages offensive? Try to judge the offensiveness of the tweets independently from your opinion but solely based on the abusive content that you may find.}}

\par{\textbf{US Presidential campaign}: \textit{The tweets in this task have been collected during the last US Presidential campaign. Would you find the content of the messages offensive?  Try to judge the offensiveness of the tweets independently from your political orientation  but solely based on the abusive content that you may find.} }

\par{\textbf{Black Lives Matter}:  \textit{These tweets are related to the Black Lives Matter protests. Would you find the content of the messages offensive? Try to judge the offensiveness of the tweets independently from your opinion but solely based on the abusive content that you may find.} }

The second part of the task description, instead, is the same for all the domains, containing a definition of what is offensive and informing the workers that there is a quality check on the answers:

\begin{table*}[h]
    \centering
    \small
    \scalebox{1}{
    \begin{tabular}{|l|r|ccc|ccc|}
    
\hline
\multirow{2}{*}{Training} & \multirow{2}{*}{Training Size} 	&	\multicolumn{3}{c|}{All domains}	&	\multicolumn{3}{c|}{Founta  + all domains}	\\
\cline{3-8}
&&	BLM	&	Covid	&	Election	&	BLM	&	Covid	&	Election \\
\hline
$A^{++}$	&	900	&	\textbf{0.756}	&	\textbf{0.752}	&	0.730	&	\textbf{0.768}	&	\textbf{0.752}	&	\textbf{0.752}	\\
$A^+$	&	900	&	0.745	&	0.724	&\textbf{	0.734}	&	0.774	&	0.736	&	0.748	\\
$A^0$	&	900	&	0.647	&	0.644	&	0.626	&	0.689	&	0.652	&	0.707	\\
\hline
$A^{++/+}$	&	1800	&	\textbf{0.776}	&	\textbf{0.756}	&	\textbf{0.732}	&	\textbf{0.779}	&	\textbf{0.738}	&	\textbf{0.750}	\\
$A^{+/0}$	&	1800	&	0.738	&	0.738	&	0.707	&	0.744	&	0.698	&	0.729	\\
$A^{++/0}$	&	1800	&	0.732	&	0.733	&	0.704	&	0.746	&	0.723	&	0.721	\\
\hline
$A^{++/+/0}$	&	2700	&	0.758	&	0.736	&	0.742	&	0.766	&	0.748	&	0.742	\\
\hline
    \end{tabular}}
    \caption{Test results on single domains using a model trained on all domains.}
    \label{tab:my_label1}
\end{table*}

\begin{table*}[h]
    \centering
    \small
    \scalebox{1}{
    \begin{tabular}{|l|r|ccc|ccc|}
    
\hline
\multirow{2}{*}{Training} & \multirow{2}{*}{Training Size} 	&	\multicolumn{3}{c|}{Out of domain}	&	\multicolumn{3}{c|}{Founta + out of domain}	\\
\cline{3-8}
&&	BLM	&	Covid	&	Election	&	BLM	&	Covid	&	Election \\
\hline
$A^{++}$	&	600	&	\textbf{0.699}	&	\textbf{0.734}	&	\textbf{0.723}	&	\textbf{0.760}	&	\textbf{0.736}	&	\textbf{0.746} \\
$A^+$	&	600	&	0.681	&	0.720	&	0.631	&	0.718	&	0.706	&	0.725 \\
$A^0$	&	600	&	0.557	&	0.603	&	0.542	&	0.674	&	0.629	&	0.672 \\
\hline
$A^{++/+}$	&	1800	&	\textbf{0.696}	&	\textbf{0.758}	&	\textbf{0.742}	&	\textbf{0.740}	&	\textbf{0.771}	&	\textbf{0.734} \\
$A^{+/0}$	&	1800	&	0.641	&	0.686	&	0.649	&	0.733	&	0.696	&	0.716 \\
$A^{++/0}$	&	1800	&	0.695	&	0.726	&	0.720	&	0.737	&	0.706	&	0.722 \\
\hline
$A^{++/+/0}$	&	2700	&	0.737	&	0.736	&	0.692	&	0.734	&	0.756	&	0.720 \\
\hline
    \end{tabular}}
    \caption{Results in the out-of-domain setting, testing the classifier on each domain when trained on the other two.}
    \label{tab:my_label2}
\end{table*}

\par{\textbf{Offensive}: \textit{Profanity, strongly impolite, rude, violent or vulgar language expressed with angry, fighting or hurtful words in order to insult or debase a targeted individual or group. This language can be derogatory on the basis of attributes such as race, religion, ethnic origin, sexual orientation, disability, or gender. Also sarcastic or humorous expressions, if they are meant to offend or hurt one or more persons, are included in this category.} }

\par{\textbf{Normal}: \textit{tweets that do not fall in the previous category.}}

\par{\textbf{Quality Check:} \textit{the HIT may contain a gold standard sentence, manually annotated by three different researchers, whose outcome is in agreement.
If that sentence is wrongly annotated by a worker, the HIT is automatically rejected.}}

Asking annotators to label the tweets independently from their views, opinions or political orientation was inspired by recent works, showing that making explicit possible biases in the annotators contributes to reduce such bias \cite{sap-etal-2019-risk}.

\section{Impact of (dis)agreement on classification - results in detail}
Table~\ref{tab:my_label1} displays domain-specific results related  to the analysis shown in Section 4.1 of the main document, where for sake of brevity we have shown only an average between the three domains. The table confirms that also on single domains, training data with higher level of agreement improve predictions, while training data with low level of agreement are  detrimental. Classification took about 2 minutes on a Titan X for the runs using only domain-specific data. Adding the intermediate fine-tuning on data from \newcite{founta2018large} increases the time to 1.5 hours.

\section{Impact of (dis)agreement on out-of-domain data - results in detail}

Similar to the previous table, Table~\ref{tab:my_label2} displays out-of-domain results related to the analysis shown in Section 4.3 of the main document, where  we report only an average between the three domains. 
The results are consistent with the average scores reported in the main document, i.e. that training data with high agreement improve prediction, while training data with low agreement are detrimental. 
Classification took about the same time of the runs in the single domain configuration.

\section{Twitter data collection}
\label{appendixd}

Through its application programming interface (API), Twitter provides access to publicly available messages upon specific request. For each of the domains analysed, a set of hashtags and keywords was identified that unequivocally characterizes the domain and is collectively  used. During a specific period of observation, all the tweets containing at least an item of this hashtags/keywords seed list were retrieved in real time (using "filter" as query). The most  relevant entries from the covid-19 seed list are: 
\textit{covid-19}, \textit{coronavirus}, \textit{ncov}, \textit{\#Wuhan}, \textit{covid19}, \textit{sarscov2} and \textit{covid}. Data were collected in the time span between 25 January and 09 November 2020.
The most  relevant entries from the blm seed list are: \textit{george floyd}, \textit{\#blm},  \textit{black lives matter}. Tweets were collected between 24 May 2020 and 16 June 2020.
The most  relevant entries from the US Elections seed list are: \textit{\#maga}, \textit{\#elections2020}, \textit{Trump}, \textit{Biden}, \textit{Harris}, \textit{Pence}.
The tweets were collected between 30 September 2020 and 04 November 2020.\\
For each domain, a big bulk of data was collected in real time for each specific time span. From these about 400,000 tweets were randomly selected and evaluated with the ensemble method as described in Section \ref{sec:data} of the main paper.

\end{document}